\documentclass[letterpaper, 10 pt, conference]{ieeeconf}  

\IEEEoverridecommandlockouts                              
\let\labelindent\relax 
\usepackage{enumitem}
\usepackage{enumitem}
\usepackage{booktabs} 
\usepackage[bookmarks=true]{hyperref}
\usepackage{tabularx}
\usepackage{graphicx}
\usepackage{lipsum}
\usepackage{epsfig}
\usepackage{array}
\usepackage{subfig}
\usepackage{cite}
\usepackage{float}
\usepackage{ifthen}
\usepackage{amsmath}
\usepackage{xcolor}
\usepackage{ntheorem}
\theoremseparator{:}

\newcommand{\fig}[1]{Fig.~\ref{#1}}

\setlist[itemize]{leftmargin=*}

\title{\LARGE \bf Human-in-the-Loop Failure Recovery with Adaptive Task Allocation}

\author{Lorena Maria Genua$^{1}$, Nikita Boguslavskii$^{1}$ and Zhi Li$^{1}$
\thanks{$^{1}$ Robotics Engineering Department, Worcester Polytechnic Institute (WPI), Worcester, MA 01609, USA {\tt\small \{lgenua,nbboguslavskii,zli11\}@wpi.edu}}
}

\begin{document}

\maketitle
\begin{abstract}

Since the recent Covid-19 pandemic, mobile manipulators and humanoid assistive robots with higher levels of autonomy have increasingly been adopted for patient care and living assistance. 
Despite advancements in autonomy, these robots often struggle to perform reliably in dynamic and unstructured environments and require human intervention to recover from failures.
Effective human-robot collaboration is essential to enable robots to receive assistance from the most competent operator, in order to reduce their workload and minimize disruptions in task execution.
In this paper, we propose an adaptive method for allocating robotic failures to human operators (ARFA).
Our proposed approach models the capabilities of human operators, and continuously updates these beliefs based on their actual performance for failure recovery.
For every failure to be resolved, a reward function calculates expected outcomes based on operator capabilities and historical data, task urgency, and current workload distribution.
The failure is then assigned to the operator with the highest expected reward.
Our simulations and user studies show that ARFA outperforms random allocation, significantly reducing robot idle time, improving overall system performance, and leading to a more distributed workload among operators.

\end{abstract}

\section{Introduction}\label{sec:intro}
Despite significant advancements in robotics, the complexity of autonomous operation in unstructured and dynamic environments makes failures in robotic systems very common and inevitable.
These robots may be deployed for healthcare (e.g., hospitals and nursing homes), public service (e.g., cafes and supermarkets), warehouses and workshops, and work for complex tasks under human supervision. 
From time to time, robots may experience \textit{technical} failures, due to hardware malfunctions or software errors, and \textit{interaction} failures, arising from uncertainties in interactions with the environments or humans in the shared workspace~\cite{honig2018resolvingfailures}.
Because these failures not only disrupt operations, but also impact human trust and influence their perception of robot reliability~\cite{liu2023failureperception, tian2021non-technical-failures, Reig2021multi-robot-perception}, it is important to have effective failure recovery strategies which can minimize idle time, enhance system performance, and mitigate adverse impacts on humans.
Although some failures can be autonomously detected and resolved by robots~\cite{shirasaka2024selfrecovery, olsen2021recovery, ahmad2024adaptablerecovery}, most failures in complex and dynamic task scenarios still require human intervention.
If the robots can request and receive assistance from bystanders, collaborators, or inexperienced users, the adaptability of the human-robot ecosystem can be enhanced significantly~\cite{honig2021hre}.
This involves identifying who to ask for help, and what type of assistance they can provide.
\begin{figure}[t]
    \centering
    \includegraphics[width=0.99\linewidth]{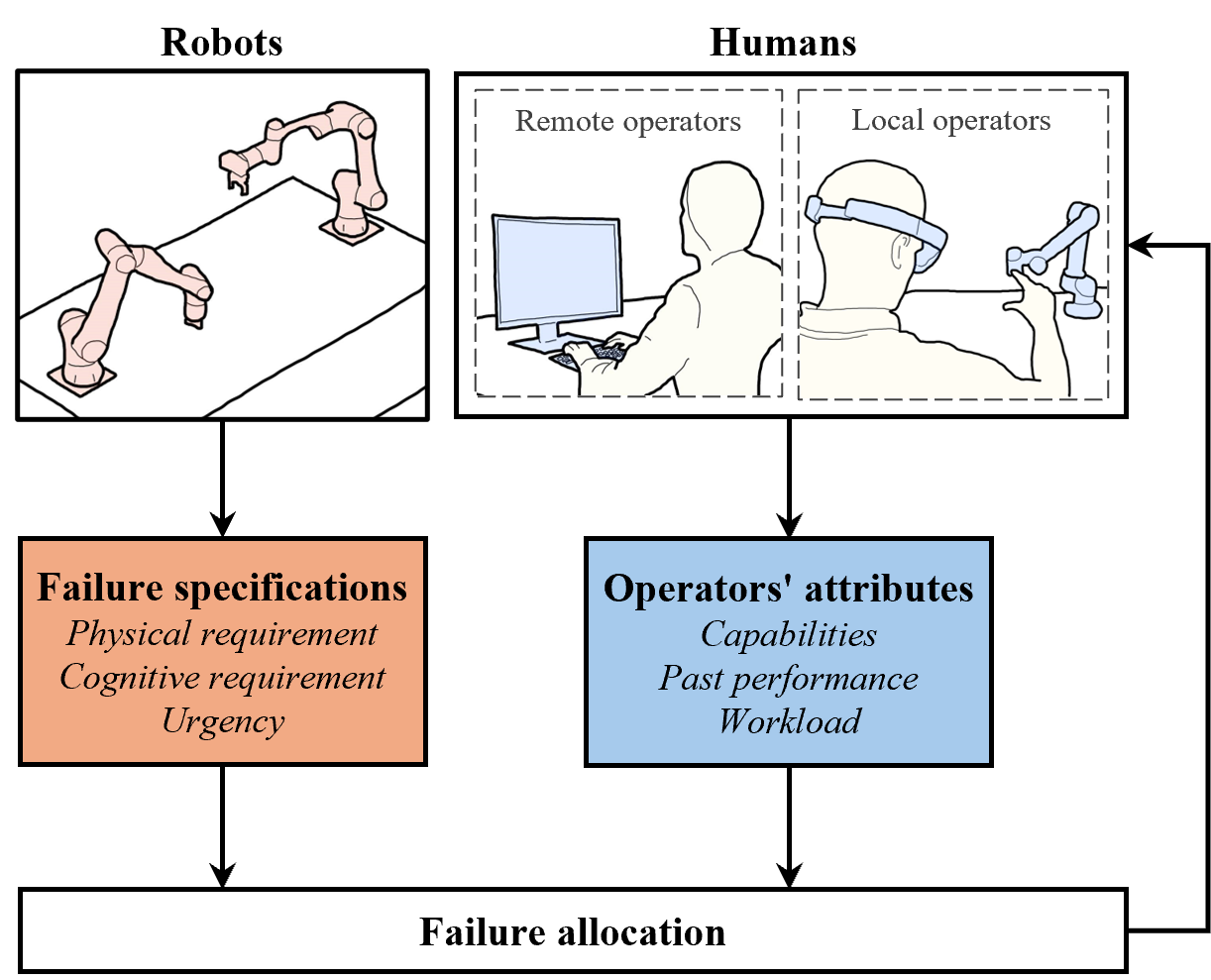}
    \caption{Our \textbf{Adaptive Robotic Failure Allocation (ARFA)} method allocates failures to the most suitable operator given the failure's requirements, the operators' capabilities, their historical performance in failure recovery, and workload distribution.}
    \label{fig:cover_picture}
    \vspace{-2ex}
\end{figure}

In our prior work~\cite{lorena2024multilateral} we developed a multilateral human-robot collaboration system, which enables robots to receive assistance from remote and local operators.
While our user study demonstrated that multilateral collaboration improves task efficiency, it also highlighted the need for dynamic and autonomous task allocation.
Failure allocation must account for each operator's skills in resolving specific issues.
For instance, a remote operator lacks the physical ability to perform certain tasks such as moving a cable from the floor or handling oversized objects the robot cannot grasp.
On the other hand, local operators, who are physically present in the robot's workspace, can address these issues quickly but may face increased workload and frequent interruptions to their primary tasks if called too often.
To this end, we propose an adaptive robotic failure allocation (ARFA) framework aimed at optimizing robotic failure recovery by optimally assigning failures to the most competent user, while distributing the workload among operators.
As shown in~\fig{fig:cover_picture}, ARFA dynamically models operator capabilities and adapts allocation based on real-time performance data and task urgency.
The results from our simulations and user study show that our proposed ARFA method significantly improves the failure recovery process and overall system performance, outperforming a baseline approach that randomly assigns failures to human collaborators.
Moreover, ARFA leads to a balanced workload between remote and local operators, considering the primary tasks they need to handle in addition to supervising and assisting the robot.  

\noindent
Our key contributions are the following:

1) We develop a framework to \textbf{enable human intervention for robotic failure recovery} with \textbf{local and remote operators}.

2) We adapt an existing task allocation strategy to a failure allocation scenario and propose a \textbf{reward function} to \textbf{reduce operators' workload} and \textbf{minimize disruptions} in task execution.

3) We validate our approach through a \textbf{user study} that simulates \textbf{local and remote operators}' interventions in \textbf{robotic failure scenarios}.

\section{Related Work}\label{sec:related}
Research in robotic failure recovery has gained substantial attention due to the increased deployment of robots in our everyday lives.
Various approaches have been explored to allow humans to best assist robots in case of failures. 
Some works have focused on generating explainable and context-specific requests for assistance, in order to facilitate communication and collaboration during the failure recovery process~\cite{knepper2015recovering, blankenburg2020recovering}.
Shared-autonomous control allows humans to oversee robotic actions and intervene with real-time corrections when needed~\cite{hagenow2021corrective}, reducing operator workload by delegating most tasks to the robot and asking for help only when needed~\cite{sankaran2021recoveryshared}.
Another promising direction is learning from demonstration, where humans teach robots to recover from failures by demonstrating recovery strategies~\cite{delduchetto2018recoverylfd, ahmad2024adaptablerecovery}.
Building on these methods to enable robots to recover and learn with human assistance, it is crucial to determine which operator is most competent to solve a failure as it occurs.
Capability-based task allocation has been studied to optimize human-robot collaboration by matching task requirements with operator and robot capabilities~\cite{lippi2021milpallocation, fabian2017allocation}.
Similarly, reinforcement learning has been applied to optimize the allocation by capturing latent dependencies between agents and tasks~\cite{Wang2023allocation}.
However, most existing methods rely on static capability assessments and offline allocation.
On the other side, while some approaches incorporate real-time monitoring to adapt allocation based on workload, they fail to model human skills~\cite{Mina2020allocation}.
These limitations make the existing methods unsuitable for allocating failures, which cannot be predicted and are time-sensitive.
\noindent
To the best of our knowledge, no existing work explicitly addresses the real-time allocation of robotic failures to human operators, particularly in scenarios involving both remote and local operators.
Our work addresses the gap in robotic failure recovery with a framework for real-time human-enabled failure recovery.
Inspired by the work by Ali et al.~\cite{ali2022trust}, which introduced a trust-based task allocation framework, we adopt the general idea of updating beliefs about agent suitability based on observed outcomes.
Binary task success or failure is used to maintain an “artificial trust” score, and tasks are allocated to maximize the expected team reward.
While Ali et al. focus on generic task allocation in a multi-agent context, their method does not generalize to failure recovery scenarios, which are time-sensitive, task-specific, and require accounting for operator workload.
Therefore, our method extends the concept to the domain of robotic failure recovery.
We explicitly model the capabilities of both remote and local human operators and continuously update these beliefs based on observed performance in failure resolution.
Our reward function is designed to guide failure assignment by considering: 1) the urgency of the failure, 2) the operator’s historical performance, and 3) the current workload distribution among operators.
This allows ARFA to assign failures to the operator whose capabilities best match the task requirements while ensuring a balanced distribution of workload.

\section{Methodology}\label{sec:method}
\begin{figure}[h!]
    \centering
    \includegraphics[width=0.98\linewidth]{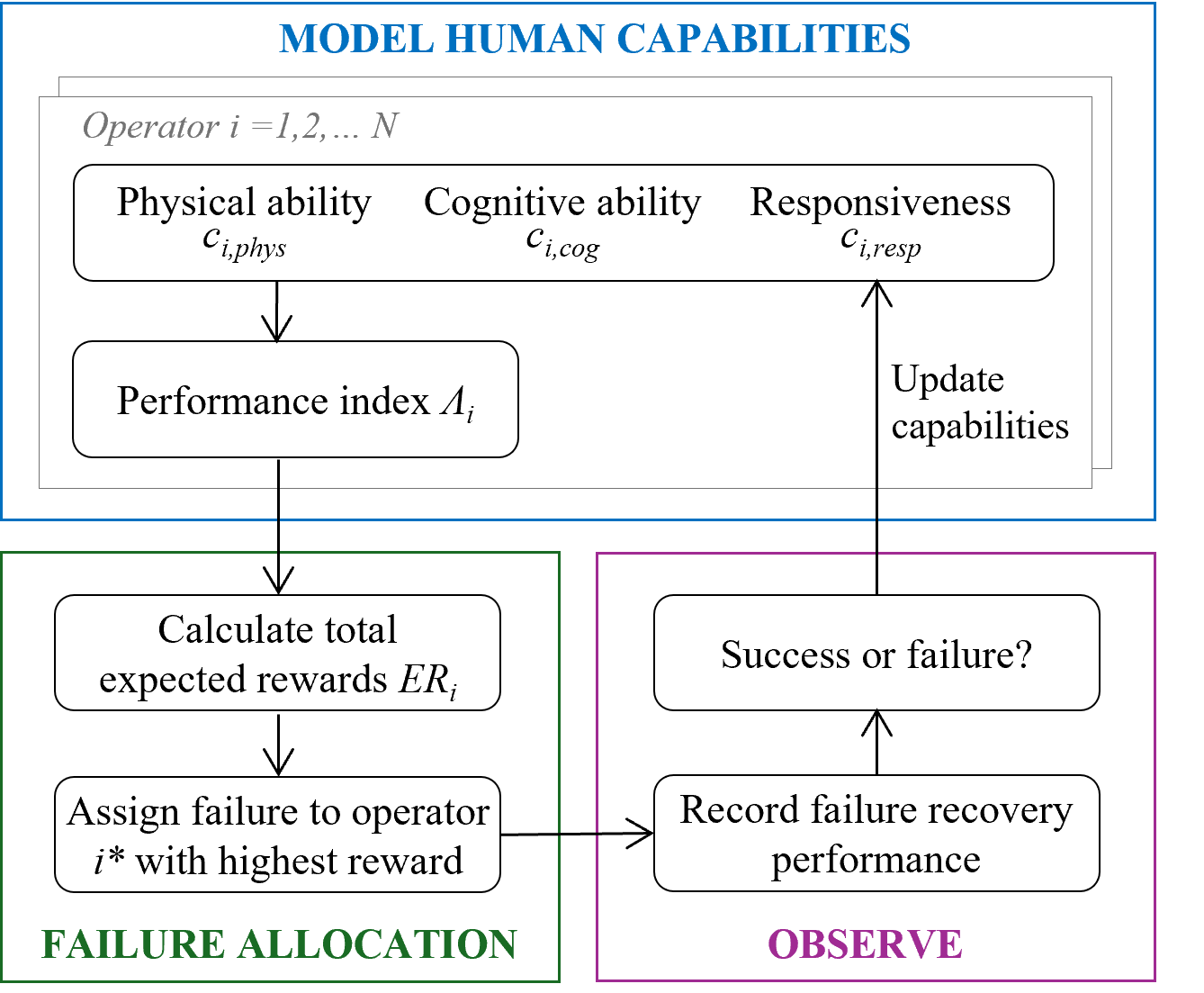}
    \caption{For a given failure, the performance index for each operator is calculated given the failure requirements and current capabilities beliefs. The failure is then allocated to the operator with the highest expected reward. Post-resolution, data is collected, and the operator's capabilities are updated based on the failure recovery outcome.}
    \label{fig:methodolofy}
    \vspace{-1ex}
\end{figure}

Our framework consists of three key components (Fig. \ref{fig:methodolofy}): first, modeling the capabilities of each operator; second, calculating the expected reward and allocating the failure; and finally, observing the operator's performance in failure recovery and updating their capabilities beliefs.

\subsection{Modeling Humans' Capabilities}\label{sec:method-model}
Let \(\mathcal{O}\) represent the set of local and remote operators. 
Each operator \( i\in\mathcal{O}\) has capabilities defined in three dimensions. These dimensions are denoted by \(\mathcal{C} = \{phys, cog, resp\} \), where \(phys\) corresponds to \textbf{physical ability}, \(cog\) to \textbf{cognitive ability}, and \(resp\) to \textbf{responsiveness}.
Responsiveness reflects the operator's ability to respond quickly and effectively to a failure, which is critical in handling urgent situations.
For each dimension \(j \in \mathcal{C}\), the capability \(c_{i,j}\) of operator \(i\) is expressed as \(\text{bel}(c_{i,j}) = (\ell_{i,j}, u_{i,j})\), where \(\ell_{i,j}\) and \( u_{i,j} \) represent the lower and upper bounds of their capabilities, respectively.
We use a uniform belief distribution as the estimation of each operator's capability in these dimensions, with 0 and 1 to be the lower and upper bounds, respectively.
This distribution captures the range within which the operator's actual capability is believed to be. 
At initialization, all bounds are set to $\ell_{i,j} = 0$ and $u_{i,j} = 1$ for each operator and capability dimension.
We further define $\mathcal{F}$ as the set of all failures encountered by the robot. 
Each failure \( f \in \mathcal{F} \) will have its own requirements in three dimensions, namely \(\mathbf{r} = [r_{\text{phys}}, r_{\text{cog}}, r_{\text{urg}}]\), corresponding to the dimensions of physical ability, cognitive ability, and urgency. 
Specifically, $r_{\text{phys}}$ and $r_{\text{cog}}$ denote the physical and cognitive workload required, respectively, for the operator to handle the failure, while $r_{\text{urg}}$ indicates how time-sensitive a failure must be handled. 
Given a failure \(f\) with requirements \(\mathbf{r}\), the capability score \(\lambda_{i,j}(\mathbf{r})\) of operator \(i\) in dimension \(j\) is calculated using:
\begin{equation}
\lambda_{i,j}(r_{j}) = 
\begin{cases} 
1 & \text{if } r_j \leq \ell_j^i \\
\frac{u_{i,j} - r_j}{u_{i,j} - \ell_j^i} & \text{if } \ell_j^i < r_j \leq u_{i,j} \\
0 & \text{if } r_j > u_{i,j}
\end{cases}
\end{equation}
Note that this approach will assign a score of 1 when the operator’s capability fully meets or exceeds the requirement, linearly decreases the score as the requirement approaches the upper bound of the operator’s capability, and assigns a score of 0 when the requirement exceeds the upper bound.
The overall performance index for every operator \(i \in \mathcal{O}\) is then computed as the weighted sum of the capability scores in all dimensions \(j \in \mathcal{C}\):

\begin{equation}
\Lambda_i = \sum_{j\in \mathcal{C}} w_j \cdot \lambda_{i,j}(r_{j})
\end{equation}
where \(w_j\) are non-negative weights that sum up to 1. The weights $w_j$ reflect the relative importance of each capability dimension and are set equally across dimensions.
\subsection{Expected Reward and Failure Assignment}\label{sec:method-reward}
We define a reward function to evaluate the assignment of a failure to a specific operator, based on the urgency of the failure and the operator’s historical performance.
A cost function accounts for the operator’s current workload, penalizing assignments that would increase their cumulative task load.

\noindent
We define reward \(R_i\) for operator \(i\) to be:
\begin{equation}  
R_i(r_{\text{urg}}, \tau_i) = \frac{r_{\text{urg}} \cdot (1 + \tau_i)}{1 + r_{\text{urg}}}
\end{equation}

where \( r_{\text{urg}} \) denotes the urgency of the failure, and \( \tau_i \) is a performance metric derived from the operator's historical failure resolution performance.
The way we define the reward will amplify differences in past performance \( \tau_i \) when the urgency \( r_{urg} \) is high, in order to increase the sensitivity to performance variations under higher urgency.
We further compute the performance metric \( \tau_i \), which indicates the operator’s past performance, by comparing the resolution times of assigned failures against a constant, task-dependent time threshold \(\epsilon\). Specifically, 
\begin{equation}
\tau_i = 1 - \frac{1}{\epsilon} \cdot \frac{\sum_{j \in \mathcal{F}}x_{ij} \cdot d_j}{\sum_{j \in \mathcal{F}}x_{ij}}
\end{equation}
where \(x_{ij}\) being a binary value indicating whether failure \(j\) is assigned to operator \(i\) (1 if assigned, else 0), and \(d_j\) represents the duration of the resolution for failure \(j\).
\\To account for the operator's workload, we subtract the cost \(C_i\):
\begin{equation}
C_i = \frac{\sum_{j \in \mathcal{F}}x_{ij} \cdot d_j}{\sum_{j \in \mathcal{F}}d_j}
\end{equation}
The cost function calculates the proportion of the total resolution time assigned to each operator, assigning a higher cost to those with a greater cumulative resolution time. 
As a result, our approach will minimize the cumulative resolution time of individual operators, ensuring a more balanced workload distribution.
Finally, we define the total expected reward \(\text{ER}_i\) as:
\begin{equation}
\text{ER}_i = \Lambda_i \cdot (R_i(r_{\text{urg}}, \tau_i) - C_i)
\end{equation}
where \(\Lambda_i\) is the performance index which takes into account how capable operator \textit{i} is at resolving failure \textit{j} with requirements \textit{r}.
The failure is then assigned to the operator with the highest expected reward:
\begin{equation}
i^* = \arg\max_{i \in \mathcal{O}} \text{ER}_i
\end{equation}

\subsection{Capabilities Optimization}\label{sec:method-optimization}
%
We further use Adam optimizer~\cite{adamoptimizer} with a learning rate and weight decay of 0.001 to update the upper and lower bounds of each capability, based on the operators' performance in resolving robotic failures.
The hyper-parameter to control the exponential decay rates of running averages was set to 0.99.
Capability bounds are updated online after each failure resolution.
The objective of the optimizer is to minimize the loss function, defined as the mean squared error between the predicted performance, computed using the current capability bounds, and the observed task success probabilities.
\begin{equation}
\mathcal{L}(\theta_i) = \sum_{\mathbf{r} \in \mathcal{R}} \left(\mathcal{P}_i(\mathbf{r}) - \hat{\mathcal{P}}_i (\mathbf{r})\right)^2
\end{equation}
where \(\mathcal{P}_i(\mathbf{r})\) is the observed success probability for operator \(i\) with failure requirements \(\mathbf{r}\), and \(\hat{\mathcal{P}}_i (\mathbf{r})\) is the predicted performance given the current capability bounds \(c_{i,j} = (\ell_{i,j}, u_{i,j})\).
After an operator is assigned to resolve a failure, the system evaluates the outcome as either a success (\(\mathcal{P}_i\) = 1) or a failure (\(\mathcal{P}_i\) = 0).
The task is considered successful if it is completed within a specified threshold, which is adjusted based on the urgency of the task.
If the resolution time exceeds this threshold, it is considered a failure.
The probability of success observed \( \mathcal{P}_i(\mathbf{r}) \) for operator \( i \) in resolving a failure with requirements \(\mathbf{r}\) is defined as the empirical probability of successful completion. 
This probability is calculated by considering the operator's performance across failures with similar requirements: 
\begin{equation}
\mathcal{P}_i(\mathbf{r}) = \frac{S_i(\mathbf{r})}{T_i(\mathbf{r})}
\end{equation}
where:
\(S_i(\mathbf{r})\) is the number of successful failure resolutions by operator \(i\) and \(T_i(\mathbf{r})\) is the total number of failure assigned to operator \(i\), with capability requirements \(\mathbf{r}\).
\\Given the current capability bounds, the predicted performance \(\hat{\mathcal{P}}_i (\mathbf{r})\) for operator \( i \) is calculated by multiplying the capability scores in all dimensions:
\begin{equation}
\hat{\mathcal{P}}_i(\mathbf{r}) = \prod_{j \in \mathcal{C}} \lambda_{i,j}(r_{j})
\end{equation}
This product assumes independence across capability dimensions; performance in one dimension does not affect the others.
\vspace{0ex}
\section{Evaluation}\label{sec:sim}
Our method was evaluated through simulations and user studies.
Our baseline for comparison is random allocation, as there is no prior work addressing failure allocation between different operators (remote and on-site).

\subsection{Simulation}\label{sec:sim-sim}
\noindent
\textbf{Environment:} We evaluated our optimized failure allocation approach (ARFA) in simulation and compared its performance with a baseline method in which failures were randomly assigned to remote and on-site operators.
We simulate one remote operator and one local operator.
Each method was tested over 20 trials, with each trial simulating 100 failures using Python 3.12.4.
The failure requirements were randomly generated, and the same were used for all trials.
%
%
For the remote operator, resolution times ranged from 40 to 90 seconds, while the local operator's times varied between 10 and 50 seconds.
This performance estimation came from the data collected during our previous user study~\cite{lorena2024multilateral}. 
A failure is considered to be successfully handled if it is resolved within a required time limit, which is dynamically adjusted based on the urgency as \({\epsilon}\) / \({(1 + r_{\text{urg}})}\),
where \( \epsilon \) is the constant time threshold set to 100 seconds, and \( r_{\text{urg}} \) represents the urgency of the failure.
\begin{figure}[t!]
    \centering
    \vspace{0.5em}    \includegraphics[width=0.99\linewidth]{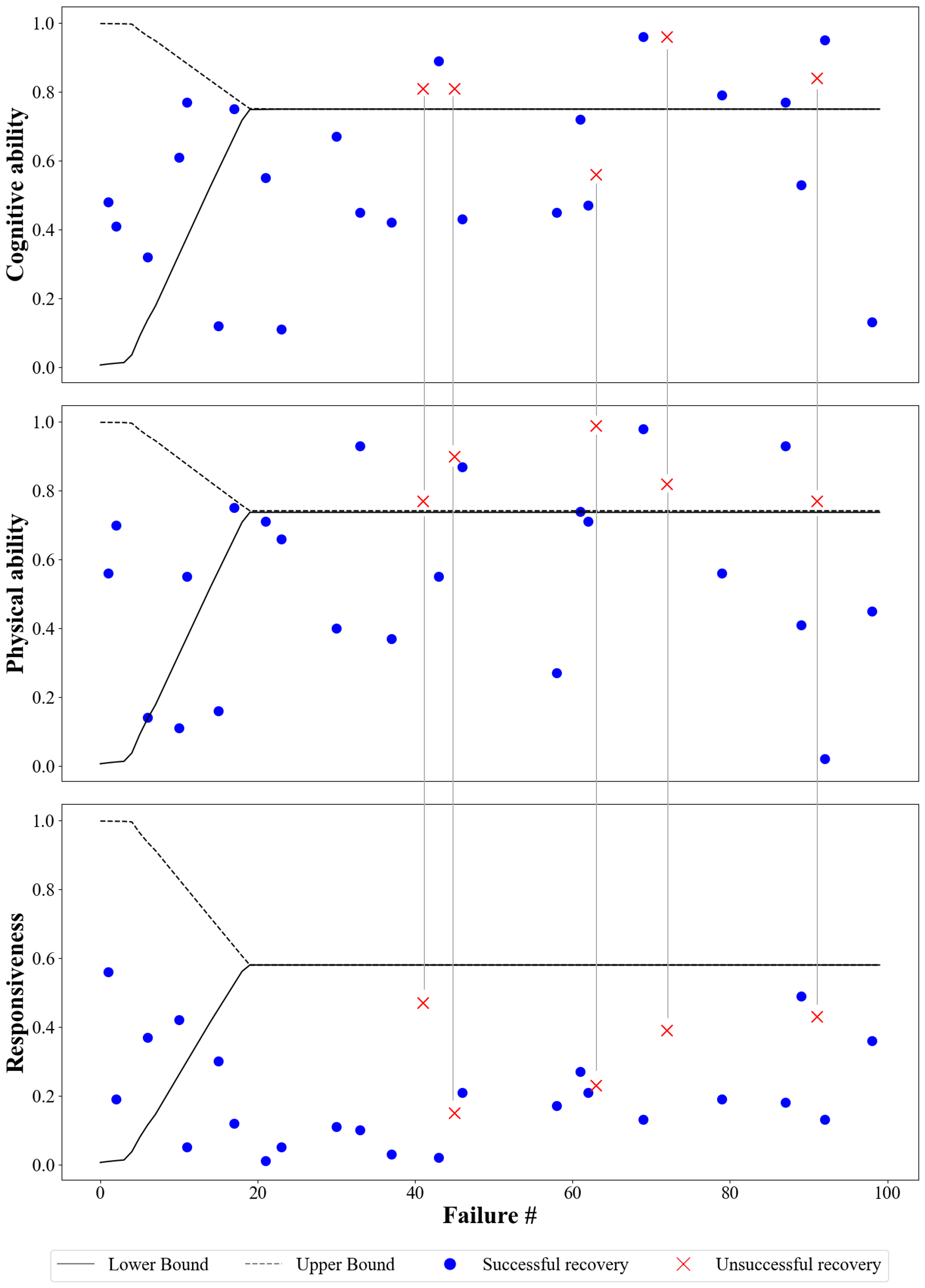}
    \caption{Simulation result of one representative trial of the remote operator. For each failure assigned to this operator, the plots show the respective \textbf{requirements} for \textit{cognitive ability}, \textit{physical ability}, and \textit{urgency}. Successfully resolved failures are marked with blue circles, while red crosses indicate unsuccessful ones. Upper and lower bounds of the \textbf{operator's beliefs for each capability} are also shown.}
    \label{fig:Capabilities}
    \vspace{-1em}
\end{figure}

\vspace{1ex}
\noindent
\textbf{Results}: Our simulation aimed to assess the optimizer's performance in updating capability beliefs and the success rate of failure recovery tasks.
Table~\ref{tab:operation_comparison} shows the success rate (mean percentage of failures resolved within the threshold time) for ARFA and random allocation.

\begin{table}[h!]
\vspace{0.5em}
\centering
\scriptsize 
\setlength{\aboverulesep}{0pt} 
\setlength{\belowrulesep}{0pt} 
\begin{tabularx}{\linewidth}{>{\bfseries}l@{}*{6}{>{\centering\arraybackslash}X}}
\toprule 
& \multicolumn{2}{c@{}}{\textbf{Local Operator}} & \multicolumn{2}{c@{}}{\textbf{Remote Operator}} & \multicolumn{2}{c}{\textbf{Team}} \\
\cmidrule(lr){2-3}\cmidrule(lr){4-5}\cmidrule(lr){6-7}
& \textbf{Mean} & \textbf{Std} & \textbf{Mean} & \textbf{Std} & \textbf{Mean} & \textbf{Std} \\
\midrule 
\textit{Random} & 100\% & 0 & 55\%\textbf{***} & 5.7 & 78\%\textbf{***} & 4.8 \\
\textit{ARFA} & 100\% & 0 & 82\%\textbf{***} & 6.8 & 95\%\textbf{***} & 1.2 \\
\bottomrule 
\end{tabularx}
\caption{Individual and Team success rates for random and ARFA allocation. (\textit{***: p$<$0.001)}}
\vspace{-1em}
\label{tab:operation_comparison}
\end{table}
\noindent
Results are reported for the local and remote operators individually, as well as their combined performance as a team.
While the local operator maintained a 100\% success rate under both Random and ARFA allocation, the remote operator had a significant performance improvement, with the success rate increasing from 55\% $\pm$ 5.7 under random allocation to 82\% $\pm$ 6.8 with ARFA (\textit{p$<$0.001}).
Similarly for team performance, the success rate increased from 78\% $\pm$ 4.8 to 95\% $\pm$ 1.2 (\textit{p$<$0.001}).

\noindent
In terms of the optimizer's performance, 98.33\% of the capabilities converged successfully across 100 trials.
\fig{fig:Capabilities} shows a representative trial for the remote operator.
This plot helps visualize, for each assigned failure, its requirements across the three dimensions—cognitive, physical, and responsiveness—and the corresponding capability beliefs for each dimension at the time it was assigned.
After each failure is resolved, the bounds are updated until convergence is reached, which occurred after approximately 20 failures in this trial.
The failures assigned try to remain below the converged capability bounds, as clearly shown for the responsiveness dimension.
Unsuccessful recoveries, marked with red crosses in the plots, typically occur when one or more requirements surpass the operator’s converged capabilities.
This happens because the reward function may prioritize distributing the workload between operators, allowing for failure assignments with requirements that exceed the operator's capabilities.
Although it may increase resolution time, it will not prevent the completion of failure resolution.
%


\subsection{User Study}\label{sec:exp}
We further conducted a user study to evaluate our dynamic failure allocation with real users.
We compared ARFA to a baseline random allocation, to answer two research questions: \textbf{(RQ1)} Does ARFA improve the failure recovery process and overall system performance? and \textbf{(RQ2)} Can ARFA balance operators' workload without compromising their primary task performance?

\begin{figure}[h!]
    \centering
    \vspace{0ex}
    \includegraphics[width=0.99\linewidth]{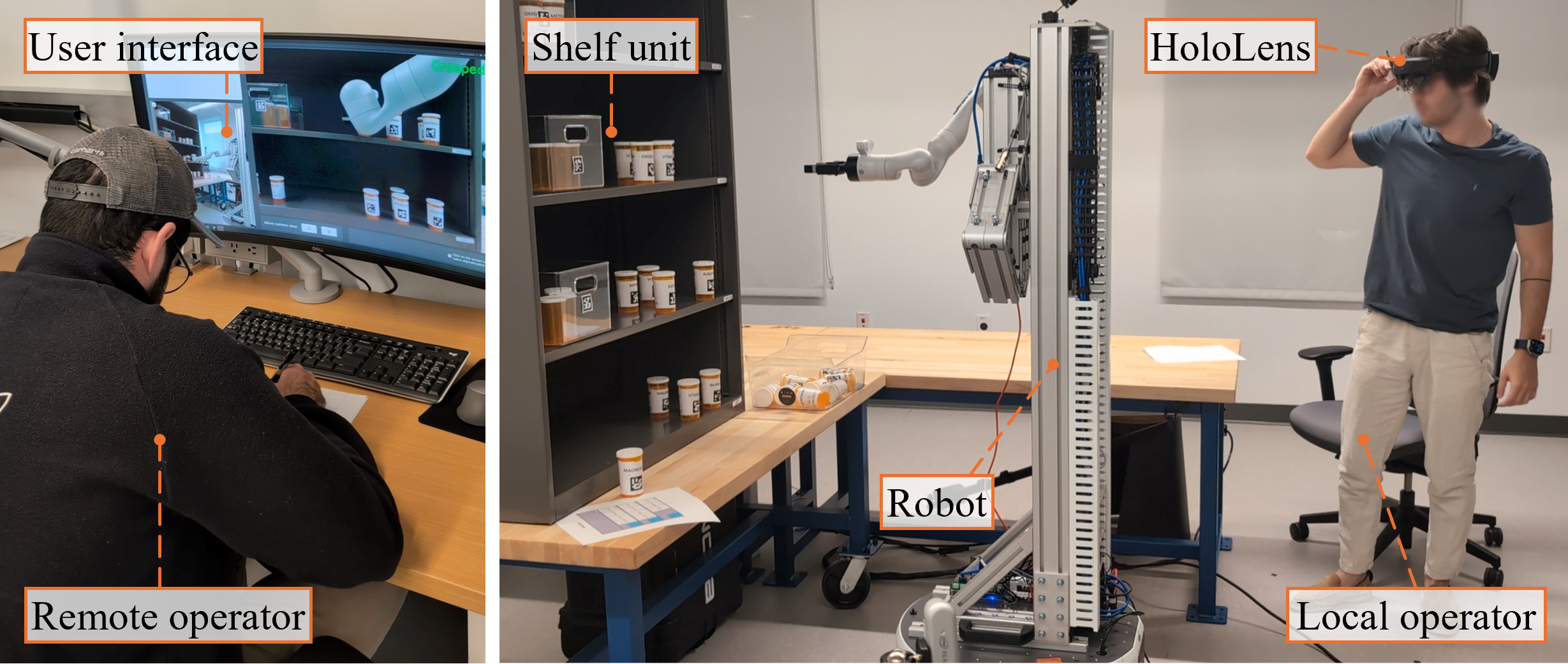}
    \caption{Experiment Setup: on the left, the remote operator uses a screen-based graphical user interface. On the right, the robotic system and the local operator wearing Microsoft HoloLens 2.}
    \label{fig:005Experiment-Setup}
    \vspace{-1ex}
\end{figure}

\vspace{1ex}
\noindent
\textbf{Experimental setup}: As shown in Fig. \ref{fig:005Experiment-Setup}, the robot used in our user study was IONA (Intelligent rObotic Nursing Assistant), the first mobile humanoid nursing assistant robot that can support multilateral human-robot collaboration~\cite{boguslavskii2023shared,lorena2024multilateral}.
IONA has a 7-DOF manipulator arm (Kinova Gen 3, with a two-fingered Robotiq 85F gripper) attached to a motorized chest that can move up and down on its mobile base (Freight Research Platform).
The robot also has two RGB+D cameras (Realsense D435) attached to its chest and arm, to provide telepresence camera views and autonomous perception.
The remote and on-site operators use augmented reality interfaces to supervise the robot's autonomous operation and assist when it encounters a failure.
\begin{table}[h!]
\vspace{-1em}
\centering
\scriptsize 
\begin{tabularx}{\linewidth}{lXccl}
\toprule 
\textbf{Failures} & \textbf{Description} & \textbf{Physical} & \textbf{Cognitive} & \textbf{Urgency} \\
\midrule 
\textit{Undetected} & The marker is not visible; the object is not detected & [0.2, 0.4] & [0.4, 0.6] & [0.0, 1.0] \\
\textit{Misplaced} & The object is occluded by another object placed in front & [0.4, 0.6] & [0.8, 1.0] & [0.0, 1.0] \\
\textit{Expired} & The object needs to be discarded and replaced with a valid one & [0.3, 0.5] & [0.2, 0.3] & [0.0, 1.0] \\
\textit{Grasp error} & Misalignment of the robot gripper results in a failed grasp attempt & [0.1, 0.3] & [0.6, 0.7] & [0.0, 1.0] \\
\textit{Non-graspable} & The object exceeds the size limit the robot can handle for grasping & [0.9, 1.0] & [0.3, 0.4] & [0.0, 1.0] \\
\bottomrule 
\end{tabularx}
\caption{Failure types with corresponding descriptions and requirements (physical, cognitive, and urgency).}
\label{tab:failure_requirements}
\end{table}
Specifically, the remote operator (shown on the left in Fig. \ref{fig:005Experiment-Setup}) uses a screen-based interface, which includes two camera views (chest and workspace camera), and buttons to send high-level commands to the robot, such as "Grasp", "Place", and "Set Aside."
The local operator uses HoloLens to render AR for displaying task and robot states, and receiving notifications for help requests from the robot.
\begin{figure*}[t!]
    \centering
    \includegraphics[width=0.95\textwidth]{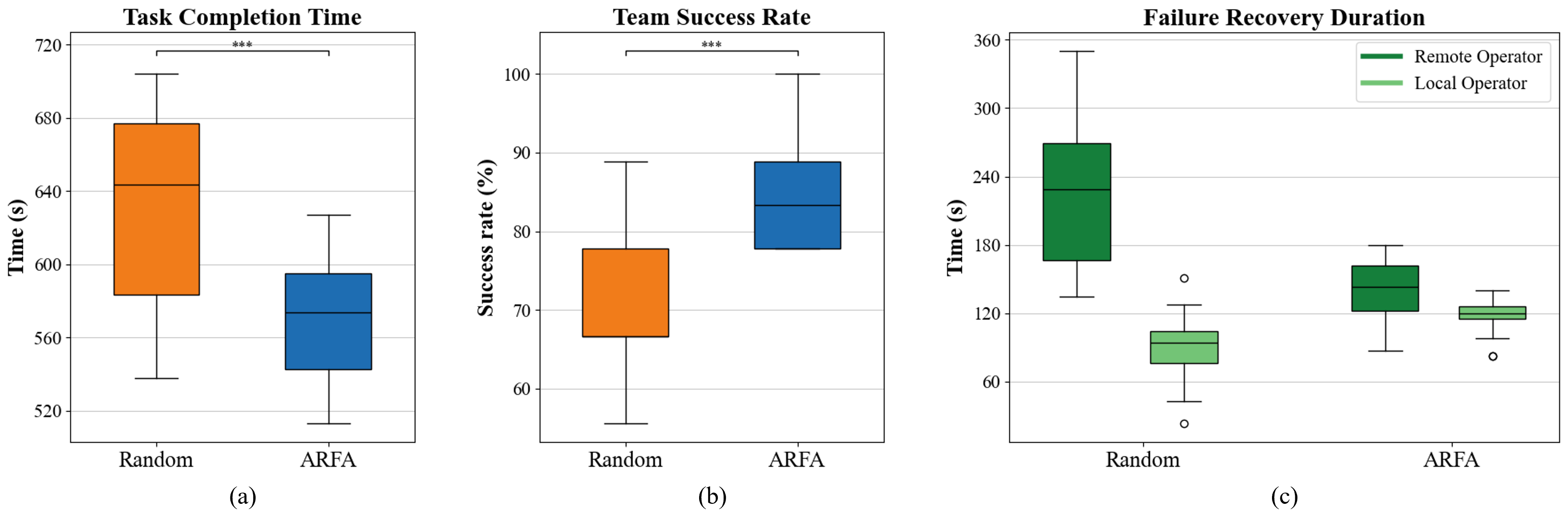}
    \caption{User study results. (a) Total \textbf{task completion time} was significantly reduced with ARFA allocation. (b) The \textbf{team success rate}, which is the ability to resolve a failure within a predefined time threshold based on the failure urgency, significantly increased with ARFA. (c) ARFA distributes the workload between operators, measured as the total \textbf{failure recovery duration} for each operator.}
    \label{fig:Results}
    \vspace{-2ex}
\end{figure*}

\vspace{1ex}
\noindent
\textbf{Task}: Our experiment simulated a nursing task in which the robot autonomously collected an order of prescribed medicines from storage shelves and placed them in a designated container, while participants were assigned the primary task of solving 4x4 Sudoku puzzles (see~\fig{fig:005Experiment-Setup}).
During the robot's operation, we simulated the failures as detailed in Table~\ref{tab:failure_requirements} along with the associated requirements that vary based on the actions necessary for operators to address them.
For example, the \textit{Non-graspable} failure has a very high physical requirement: users have to manually move objects that the robot cannot grasp, and therefore can only be resolved by local operators.
Meanwhile, the \textit{Misplaced error} has a higher cognitive demand, because the operator needs to recognize that the medicine was not detected due to occlusion and to determine the correct location for the misplaced item.
When such failures occurred, the robot would allocate the failure to one participant, who had to pause the primary task to assist the robot.

\vspace{1ex}
\noindent
\textbf{Participants and procedure}: 
Our user study recruited N=20 participants (14 males and 6 females, 25.3 $\pm$ 3.42 years).
The study was approved by the Worcester Polytechnic Institute IRB on May 8, 2024.
Each user study involved 2 participants, designated as the remote and local operators.
The experiment consisted of three phases: 1) Training, 2) Data Acquisition, and 3) Performance.
During the \textbf{Training} phase, the experimenter provided an overview of the HoloLens and screen-based interfaces, introduced the potential failure types they will encounter, and instructed participants on how to address them.
Then, participants practiced resolving each type of failure at least once, continuing until they reported feeling confident in operating the interface.
In the \textbf{Data Acquisition} phase, participants repeatedly handled failures for 20 minutes.
Here, no specific allocation method was used; instead, the robot assigned each failure to participants in an alternating sequence.
In this phase we gather data on operator performance while handling failures with varying requirements. 
Following each failure resolution, performance outcomes were fed to the optimizer to update the capability bound beliefs of each operator. 
The final updated capability bounds at the end of the Data Acquisition phase were stored and subsequently used as input for the dynamic assignment method in the \textbf{Evaluation} phase.
During the \textbf{Evaluation} phase, we conducted two trials for each mode (Random and Dynamic assignment), and the order was randomized for each participant.
%

\vspace{1ex}

\noindent
\textbf{Results}: We used the non-parametric Wilcoxon signed-rank test to compare ARFA and Random allocation.

\textbf{(RQ1) System performance:} To evaluate the reduction in robot idle time—the period during which the robot remains in a fault state awaiting resolution. We measured the \textit{total duration of failures}, defined as the time elapsed from when a failure is communicated to an operator until it is resolved.
Our results show that resolution time significantly decreased with ARFA allocation compared to Random allocation (\textit{p$<$0.001}). 
Specifically, with Random allocation, the mean duration of failures was 312.73 s $\pm$ 52.1, while ARFA reduced it to 257.6 s $\pm$ 33.99.\\
We also analyzed the \textit{task completion time}, the team and individual success rate, and the operators' primary task performance. 
The significant reduction in failure resolution times contributes to a decrease in overall task completion time, which includes the robot's autonomous operations and failures.
Indeed, ARFA allocation resulted in significantly faster task completion times, with a mean of 571.85s $\pm$ 33.95 compared to Random allocation, which had a mean of 632.98s $\pm$ 54.47 (\textit{p$<$0.001}) (\fig{fig:Results} (a)).
\\~\fig{fig:Results} (b) shows the \textit{team success rate}, defined as the ability to resolve a failure within a predefined time threshold.
ARFA significantly improved the team’s success rate, increasing it from 69\% $\pm$ 9 under Random allocation to 85\% $\pm$ 8 (\textit{p$<$0.001}).
We further examined \textit{success rates separately for remote and local operators}.
Remote operators had a significant improvement under ARFA, achieving a mean success rate of 55\% $\pm$ 23 compared to 40.5\% $\pm$ 14 under Random allocation (\textit{p$<$0.01}).
For local operators, they were consistently high for both allocation methods, with no significant difference observed (\textit{p$>$0.05}).
Local operators achieved a perfect success rate of 99.2\% $\pm$ 3.7 under Random allocation and 100\% under ARFA. 

\textbf{(RQ2) Workload:} We evaluated workload distribution by analyzing the time each operator spent on failure recovery.~\fig{fig:Results} (c) shows that ARFA reduced the mean and variance of the remote operator’s recovery times and narrowed the disparity between the two operators.
Specifically, the difference in recovery times between the two operators was 132.30s $\pm$ 79.69  under Random allocation, which ARFA reduced to 33.64s $\pm$ 14.47 (\textit{p$<$0.001}).
Primary task performance, measured as the number of Sudokus solved per minute, remained consistent between Random (1.38 $\pm$ 0.53) and ARFA (1.37 $\pm$ 0.57) modes and showed no significant difference (\textit{p$>$0.05}).%
%

\section{Conclusion}\label{sec:con}
In this paper, we bridge the gap in real-time robotic failure recovery by proposing ARFA, an adaptive failure allocation framework that dynamically assigns robotic failures to the most suitable human operator based on their capabilities, historical performance, and workload distribution.
Through simulations and user studies, we demonstrated that ARFA significantly enhances failure recovery performance by reducing robot idle time, improving success rates, and promoting a balanced workload among operators.
Despite these positive results, ARFA can still be enhanced to address certain limitations.
The framework relies on simplified models of human capabilities, which do not fully capture context-dependent, nonlinear factors such as fatigue and cognitive loads.
It also assumes that all failures are fully characterized and belong to a predefined set, with requirements aligned along three capability dimensions.
Additionally, noisy data from remote operators can lead to sparse gradients, making it harder for the responsiveness capability to converge.
Our future work will focus on adapting the allocation for uncontrolled scenarios with more types of failures.
Additionally, we will evaluate the generalizability of the framework with larger teams of operators.

\bibliographystyle{IEEEtran}
\bibliography{references}

\end{document}